# Bayesian Learning of Causal Relationships for System Reliability

Xuewen Yu[1], Jim Q, Smith[2] and Linda Nichols[3]

[1,2,3]*Statistics Department, University of Warwick, United Kingdom.*
[2]*The Alan Turing Institute, United Kingdom.*
[1]*xuewen.yu@warwick.ac.uk*
[2]*j.q.smith@warwick.ac.uk*
[3]*l.nichols@warwick.ac.uk*

Causal theory is now widely developed with many applications to medicine and public health. However within the discipline of reliability, although causation is a key concept in this field, there has been much less theoretical attention. In this paper, we will demonstrate how some aspects of established causal methodology can be translated via trees, and more specifically chain event graphs, into domain of reliability theory to help the probability modeling of failures. We further show how various domain specific concepts of causality particular to reliability can be imported into more generic causal algebras and so demonstrate how these disciplines can inform each other. This paper is informed by a detailed analysis of maintenance records associated with a large electrical distribution company. Causal hypotheses embedded within these natural language texts are extracted and analyzed using the new graphical framework we introduced here.

## 1   Introduction

The Bayesian Network (BN) has proved to be one of several efficient machine learning graphical tools for modelling conditional independencies and making inference about relationships between many variables. Concurrently advances in causal modelling has led to a better understanding of how to predict complex systems when these are subjected to control. A major breakthrough then occurred about 20 years ago when it was then discovered that causal and graphical modelling could be combined. This paper reports on how the combination of these two technologies are being applied to give more insights concerning causal hypotheses that relate specially to reliability and system safety.

Of course causal ideas have been embedded within reliability theory for a long time, both to explore the reasons behind a failure and to also estimate the efficacy of various interventions in the system that might ameliorate adverse outcomes. However, the graphical frameworks around which these ideas appear have been tree like – for example fault trees or event chains (Leverson 2011) – rather than the most common graphical framework for analyzing causation: the BN. Only recently have graphical causal methods emerged for methodologies and algorithms to exist for causal discovery and causal reasoning on such tree structures. The primary such graph is the chain event graph (CEG), see Thwaites, Smith and Riccomagno (2010), Barclay, Hutton and Smith (2013) and Collazo *et. al* (2018). This class of probabilistic graphical model enables us to search for and then hypothesize causal relations between events and to estimate effects of external interventions via a Bayesian hierarchical learning process, transferring technologies originally designed for BNs. Here we harness this CEG technology transfer to develop new causal methodologies to shine a new light on causal analysis within reliability theory.

In this paper we import the concepts of "root causes" and "remedial work", that are currently central ideas within reliability, into a graphically based causal model and relate these to engineers' written explanations. A "root cause" of a failure is the initial contributing factor that leads to this defect and is the target of an ideal "remedial work" after a failure is to fix this root cause. In reliability theory, a root cause analysis is usually supported by a fault tree analysis where "bottom level" nodes in the tree depict root causes (Gano 2011). In contrast, the model framework we develop here is described by trees that respect the chronological order in which they might happen, and so, in particular, the causal story behind the fault. This alternative representation then enables us to capture various causal hypotheses as fragments of these CEGs. They provide the vehicle through which reported engineering expert judgements can be systematically embedded into a statistical analysis of failure data. These causal fragments are first extracted from the natural texts obtained from a maintenance log. We then demonstrate how a Bayesian learning algorithm constructed from these extracted explanations can provide automatically generated inferential support about the root causes of failures. We argue that, if properly implemented, such methods could provide powerful decision tools for an automized causal fault analyses which, most importantly, can be scaled up. Expert judgements are captured by both the hypothesized framework depicting how faults might occur and prior probabilities about the likelihood of each step in these paths to failure.

Standard BN causal model would classify "remedial work" as a particular portfolio of an external intervention (Pearl 2000, Iung *et.al* 2005). Here instead we use the more recently developed CEG causal algebra (Thwaites, Smith and Riccomagno 2010) to express this intervention. In order to do this, a remedial intervention needs to be carefully defined. In particular, we need to define a new "*do*" operation that makes it possible for a graphically supported causal analysis to apply to intervened system in reliability studies.

In Section 2 we review the concept of a CEG. Then in Section 3 we give a detailed explanation of how we define a remedial intervention and show how to mathematically express such intervention in terms of causal algebras and systematically reconstruct a CEG after a remedial intervention. This is followed by a simulation study in Section 4 demonstrating the efficacy of adding causal information into a reliability analysis.

## 2   Chain Event Graph and Causal Concepts

Here we develop a methodology to automatically accommodate *all* information embedded within maintenance logs of plants into the analysis of failure data. Such logs consist of numeric and categorical data, such as time and failure mode, but are also supplemented by natural texts written by engineers explaining what and why they have chosen certain remedial actions. These texts can obviously be extremely informative about the causes, symptoms, defects that have occurred (Iung *et.al* 2005). A graphical framework is a useful tool to accommodate this information. We argue here that a tree based framework such as a CEG rather than a BN network should be the preferred such graph. So we next illustrate how to represent the explanations embedded within these texts to construct hypothesized or chains of events that might have led to a failure within a CEG.

So let an event tree $T = (E_T, V_T)$ have an associated probability vector $\boldsymbol{\theta}_T = (\boldsymbol{\theta}_v)_{v \in V_T}$, where each edge emanating from $v \in V_T$ is annotated by a component of $\boldsymbol{\theta}_v$, each component of $\boldsymbol{\theta}_v$ being associated with one of the edges out of $v$. A staged tree is a colored tree that embellishes an underlying event tree. Two vertices $v$ and $w$ are said to be at the same stage, given the same color, if $\boldsymbol{\theta}_v = \boldsymbol{\theta}_w$ up to a permutation of their components (Collazo *et.al* 2018). Edges out of $v$ and $w$ are then colored the same if they correspond to events assigned the same probability. A new graph, the CEG, is then defined as follows.

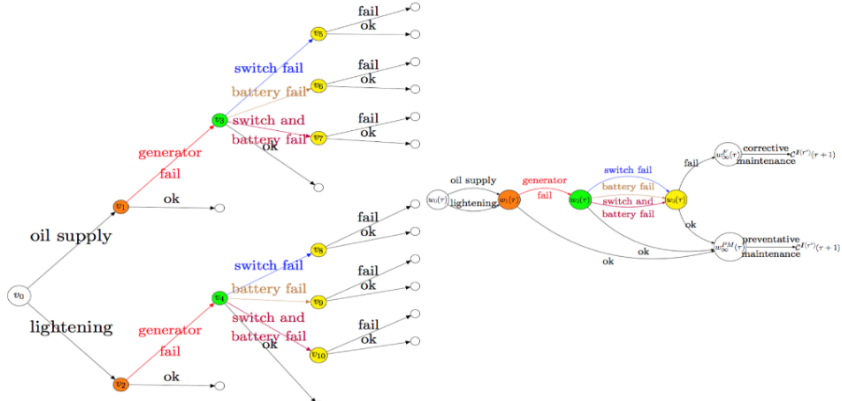

Figure 1. Example of a staged tree for security system is shown on the left, the event tree structure is elicited from the fault tree and description of it in Bedford and Cooke (2001); An example of the dynamic process on CEG is shown on the right.

**Definition 1 (Chain Event Graph)** *A Chain Event Graph (CEG) $C(T) = (W, F)$ is a graph with vertex set $W = W_T$ given by the set of positions in the underlying stage tree. A position is the set of stages whose event subtrees have the same colored graph and the same associated sets of edge labels. Every position inherits its color from the staged tree. F is a set of edges between these vertices with the following properties. If there exist edges $e = (v, v'), e' = (w, w') \in E$ and $v, w$ are in the same position, then there exist corresponding edges $f, f' \in F$. If also $v', w'$ are in the same position, then $f = f'$. The labels $\theta(f)$ of edges $f \in F$ in the new graph are inherited from the corresponding edges $e \in E$ in the staged tree. Then $(C(T), \boldsymbol{\theta}_T)$ is called a Chain Event Graph (Barclay et.al 2015, Collazo et.al 2018).*

A causal CEG orders events along its root-to-sink paths to be consistent with any hypothesized temporal ordering of events. This is especially useful when events that might happen after a control – here a remedial action – are expressing various causal hypotheses. We will see that a remedial action will only be able to affect the train of events that happen after the intervention, *i.e.* downstream from the root of the CEG. An example of such a CEG is given in Figure 1, which starts with root causes: {*oil supply, lightening*}. The CEG of a reliability system starts with partitions of components and failure modes followed by paths describing putative causes whilst symptoms are represented towards the sink of the CEG. An indicator variable defined at the end of each path indicates whether that part has failed.

## 3 External Intervention

We next devise a bespoke intervention calculus analogous to the calculus of Pearl (2000) for BNs. This embellishes the tree based CEG calculus in Collazo *et al.* (2018) to make this suitable to the domain of reliability. Here for brevity we will focus our discussion on expressing graphically the central concept of a root cause and their potential remedies. This provides us with a framework within which natural language explanations in maintenance logs describing faults and remedial actions taken can be systematically embedded into particular families of probabilistic models subjected to control. For a discussion of the practicalities of how the expert judgements we refer to below are elicited, how data can be used to refine these estimates in non-vanilla examples and how NLP methods can be used to extract the causal fragments see Yu *et al.* (2020).

## 3.1 *A classification and new semantic for remedies*

Previous work (Iung *et.al* 2005, Borgia *et.al* 2009, Cai *et.al* 2013) classified maintenance into "perfect/minimal/imperfect maintenance". Here we use terminologies closely parallel previous ones. If an intervention successfully addressed the root cause of a fault, it will be called a "remedy". Under this intervention we will assume that all subcomponents associated to the root cause of the failure are "renewed" and the status of the remedied part is "as good as new" ("AGAN") (Bedford and Cooke 2001, Andrzejczak 2015). Let $R(r^*)$ denote the result of a remedial intervention $r^*$ that observed from maintenance logs. The space of $R(r^*)$ is partitioned $\mathcal{R}^* = \{R^P, R^{IP}, R^U\}$ so that classifies the results into three types.

A remedial act $r^*$ is a **perfect remedy** - *i.e.* in $R^P$ - when the root causes of the failure are fully addressed and corrected by $r^*$. A graphical illustration for this type is shown in Figure 2(a). The dashed directed edge connecting from the last symptom to the root node represents the status of the corresponding piece of machine being returned to just before any defect occurs: an "AGAN" status, coinciding with "perfect maintenance" as defined by Iung *et.al* (2003).

When the root cause is not remedied by $r^*$, but only a subset of the secondary or intermediate faults are remedied, this is an imperfect result $R^{IP}$, $r^*$ is termed as an **imperfect remedy**. The condition of the defect equipment is returned to the status after occurrence of root cause, which is not "AGAN", see (b) where direct dashed lines pointing from the sink node to the vertices on the downstream path of the edge labelled as root cause. Here to remedy the root cause additional maintenance will be required. This is represented by the grey vertex in (b). When from logs the result is uncertain immediately after $r^*$, it is classified as $R^U$ – an **uncertain remedy**. Diagnostic information is not yet available so the root cause of the failure cannot yet be determined. As yet unknown and undescribed follow-up maintenance might be needed: see (c).

To define this graphical construction more generically, we next define a new variable to indicate whether a root cause is targeted by a remedial intervention. This is a binary vector used to indicate which root causes are targeted by the remedial action.

**Definition 2 (Intervention Indicator)** *Let $I(r^*)$ denote the intervention indicator defined for remedy $r^*$ for all root causes $E(root)$ with length $|E(root)|$. Let $I(r^*) \triangleq \{I_{e_1}(r^*), \ldots, I_{e_{|E(root)|}}(r^*)\}$ for edges $e_i \in E(root)$, $E(r^*)$ denote the set of edges intervened on when applying remedy. So,*

$$I_{e_i}(r^*) = \begin{cases} 1, & \text{if } e_i \in E(r^*) \\ 0, & \text{otherwise} \end{cases} \quad (1)$$

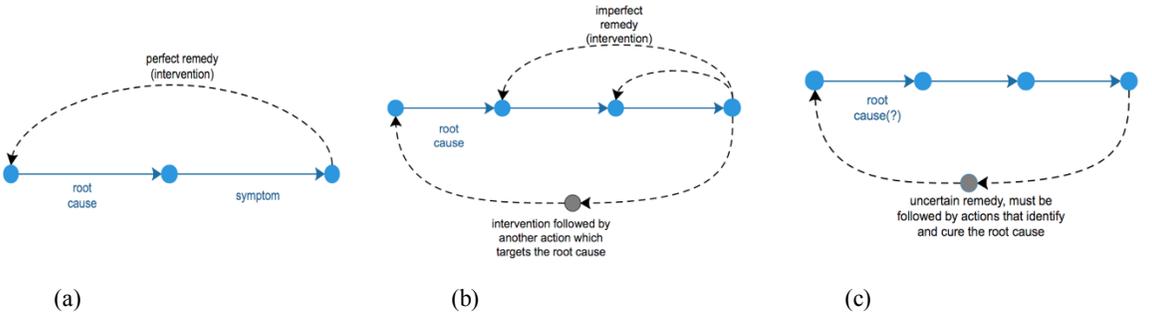

(a)　　　　　　　　　　(b)　　　　　　　　　　(c)

Figure 2. Different types of remedies

### 3.2 *Causal algebra for remedial interventions*

Adapting Pearl's "*do*"-operation for use with a CEG is given in Collazo et.al (2018). Retain the notation of Section 2 used for a CEG in an "idle" system – i.e. one not under any control. Now for a manipulated CEG $\hat{C}$, where we have forced an event labelled on edge $\hat{e}$ to happen, the conditional probability $\theta(\hat{e}) = 1$; and all other edges $e_i$ emanating from the same vertex as $\hat{e}$ have conditional probability $\theta(e_i) = 0$.

Here we now need to introduce a new the intervention on CEG, designed specifically for the analysis of fault data as of the type we illustrated in Section 3.1. We first illustrate the notation for two specific types of events: a failure event and its root cause event. In the idle CEG denote the probability of failure by $\pi(\Lambda(fail)|T)$ where $\Lambda(fail)$ denotes all paths passing edges labeled as "fail". Let $\pi(E(root)|T)$, $E(root) \in E(T)$, denote the edge probabilities associated with root causes in the tree. Partition the paths $\Lambda(fail)$ by elements of $E(root)$.

After an external intervention $r^*$, unlike standard causal algebras, here the manipulated tree preserves the topology of $T$. We perform a stochastic manipulation on the CEG by reassigning the probability distribution over root causes: $\pi(E(root)|T)$. The probability of failure given the external intervention $r^*$ is now:

$$\pi^*(\Lambda(fail)|do(r^*), T) = \frac{\pi(\Lambda(fail)|T)}{\pi(E(root)|T)} \pi^*(E(root)|do(r^*), T) \qquad (2)$$

This stochastic manipulation over root causes is then implemented by importing $I(r^*)$ into the system. The new probability $\pi^*(E(root)|I(r^*), T)$ is specified externally through domain expert judgements. Considering the uncertainties of $I(r^*)$, the new probability distribution over root causes after intervention has taken place can be expressed:

$$\pi^*(E(root)|do(r^*), T) = \sum_{I(r^*)} \pi^*(E(root)|I(r^*), T) p(I(r^*)|do(r^*), T) \qquad (3)$$

The indicator $I(r^*)$ is the tool that delivers the effect of intervention. Depending on the types of intervention $I(r^*)$ could be directly observed or have a distribution. Thus:

(i) Under perfect intervention:
$$p(I(r^*) = I^P(r^*)|do(r^*), T) = 1 \qquad (4)$$

(ii) Under imperfect intervention:
$$p(I^{IP}(r^*)|do(r^*), T) = \int_0^1 \sum_{x^a \in X^A(r^*)} p(I^{IP}(r^*)|\gamma) f(\gamma|x^a, r^*) p(x^a|do(r^*)))\, d\gamma \qquad (5)$$

(iii) Under uncertain intervention:
$$p(I^U(r^*)|do(r^*), T) = \int_0^1 \sum_{x^a \in X^A(r^*)} p(I^U(r^*)|\gamma) f(\gamma|x^a) p(x^a|do(r^*)))\, d\gamma \qquad (6)$$

In the above equations, $x^a \in X^A(r^*)$ is the specific action taken following $r^*$, where $X^A(r^*)$ can be elicited from a domain expert. Here $\gamma = \{\gamma_{e^r}\}_{e^r \in E(root)}$, where $\gamma_{e^r}$ denotes the probability of the root cause labelled on $e^r \in E(root)$ being remedied.

### 3.3 *Causal algebras and the analysis of the impact of interventions in a CEG*

A semi-Markov process defined on the CEG is now used to model the failures and maintenance activities described within a log. The state defined in semi-Markov process is analogous to the state of machines represented on the tree. So represent the renewal kernel by $Q_{ij}(t)$, with transition probabilities $\theta_{ij}$ and holding times distribution $f_{ij}(t; \boldsymbol{\mu}_{ij})$:

$$Q_{ij}(t) = \theta_{ij} f_{ij}(t; \boldsymbol{\mu}_{ij}) \qquad (7)$$

The effect of an intervention is to assign the new probability distributions of a subset of root causes. Here both the transition probabilities and the holding times distributions associated to edges of the CEG can be a function of the chosen manipulation. Once the root cause labelled on $e_r$ is remedied it is less likely the next failure is still caused by it. The transition probability along this edge typically reduces whilst its associated holding time is expected to increase.

To express this mathematically we next define explicit maps to update the distribution after control. Let $\boldsymbol{\alpha}^*_v$ and $\boldsymbol{\alpha}_v$ be the new and old hyper parameter vectors after manipulation for transition probability vector $\boldsymbol{\theta}_v$. Define the map $g$ updating $\boldsymbol{\alpha}^*_v$ from $\boldsymbol{\alpha}_v$ as

$$g: (\boldsymbol{\alpha}_v, \boldsymbol{I}(r^*), \omega_v) \mapsto \boldsymbol{\alpha}^*_v \quad (8)$$

Here the parameter $\omega_v > 0$ controls the effect of $\boldsymbol{I}(r^*)$ on $\boldsymbol{\alpha}_v$. Assume $\omega_v$ is either a known value, set by experts, or has a known distribution with hyper-parameters again whose values are set by experts. We assign a probability distribution to the intervention indicator $\boldsymbol{I}(r^*)$: if $r^*$ is not perfect we learn this through a Bayesian model. The distributions and the associated parameters can be determined by domain experts. We allow flexibility in the choice of the transformation $g$ given different circumstances. The exact form of $g$ depends on expert judgements and any relevant past data.

Let parameter $\beta_v > 0$ represent the strength of an intervention on the holding time distributions for edges emanating from the vertex $v$. This is set by users or has a known distribution. Similarly, we define a function $J$ such that:

$$J: (\boldsymbol{\eta}_v, \boldsymbol{I}(r^*), \beta_v) \mapsto \boldsymbol{\eta}^*_v \quad (9)$$

where $\boldsymbol{\eta}_v$ and $\boldsymbol{\eta}^*_v$ are the vectors of hyper-parameters for the holding time distributions over all edges emanating from $v$ before and after manipulation respectively.

The methodology for choosing the precise form of these functions is beyond the scope of this paper. However, detailed explanations and examples for how to choose these functions in a variety of circumstances based on natural language texts are given in Yu *et.al* (2020).

## 4 Experiment

In this section we will show the potential additional value of embedding causal elements extracted from natural language texts into fault analysis using a simulation study. In particular, we demonstrate how estimation and prediction improves when such information is not ignored.

Here we use our revised framework to compare the estimation of parameters on CEG that formally takes account of the applied intervention (for details see supplementary materials) with a model that does not using an artificial dataset when we have a perfect description of the system. This helps bound the advantages we can expect from utilizing natural language texts. Assume the ground truth tree shown in Figure 3(a) and also the stages and positions are known. Nodes in the same stage are colored in orange: $\{v_2, v_3\}$, green: $\{v_5, v_6\}$ and pink: $\{v_8, v_9\}$ respectively. There are clusters of edges that share the same holding time distributions: $\{e_{24}, e_{36}\}$, $\{e_{25}, e_{37}\}$, $\{e_{58}, e_{69}\}$, $\{e_{81}, e_{91}\}$, $\{e_{82}, e_{92}\}$. For the data generating process, we set up ground truth transition probabilities and time for edges in the tree without specifying them separately for the intervened and non-intervened system. These parameters over edges can be estimated via the algorithm for either of the two regimes. We generated 10 groups of machines' data such that in each group, where these machines are assumed to have the same maintenance history and share the same parameters for conditional probabilities and holding time distributions. In total, we generated 5 different dataset with sizes 500, 1000, 3000, 5000 and 10000 respectively.

Assume transition probabilities are $\boldsymbol{\theta}_v \sim Dirichlet(\boldsymbol{\alpha}_v)$ and their holding time is drawn from $h_e \sim Weibull(\xi_e, \eta_e)$. Prior elicitation for $\boldsymbol{\alpha}_v, \xi_e, \eta_e$ is discussed in Yu *et.al* (2020). An MCMC-within-Gibbs algorithm enabled parameters to be estimated. Agglomerative Hierarchical Clustering then selected a CEG representation for the data. Three levels of stages could logically be merged: $l_1 = (v_2, v_3)$, $l_2 = (v_5, v_6)$, $l_3 = (v_8, v_9)$. We ran 100 simulations for different numbers of phantom units for each sample size. Figure 3(b) presents the proportions of the simulated results that merge two stages in the same level or two edges in the same cluster. We can observe the intervened model agrees with the ground truth in terms of merge of stages. On the other hand, non-intervened model cannot estimate the CEG structure well, especially for stages in $l_2$ and $l_3$ even with a large sample size. It is also evident that with a sample size greater than 3000, the merging of edges in terms of the holding time distributions can be well estimated by the algorithm when our intervention is acknowledged. The advantages of incorporating causal factors stand out especially for larger dataset when comparing the estimates with ground truth: see Fig. 4. For example, for 5000 samples, the mean cluster error is 0.31 for intervened model, 0.116 smaller than the non- intervened model. We also compared the posterior estimate of mean holding times and the real ones. The intervened model has mean error 0.28 while the non-intervened gives 0.32. The prediction in terms of transition probabilities is also improved by 0.16.

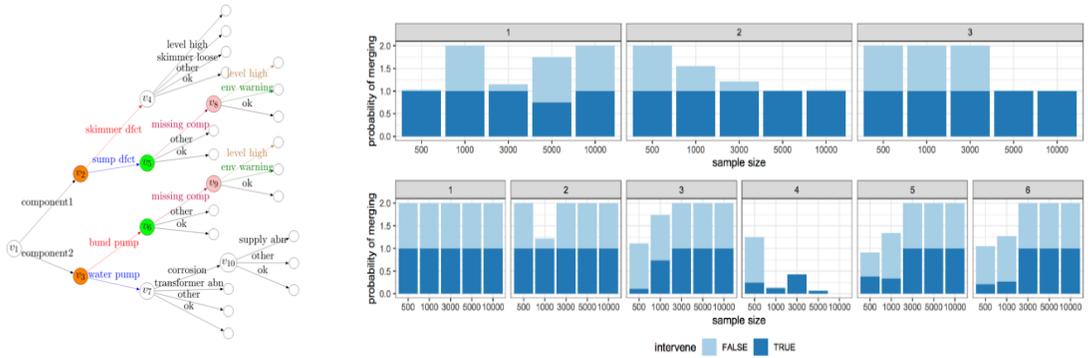

(a)                                                                                 (b)

Figure 3. (a) Event tree used for simulation study; (b) Comparison of models with and without intervention: the top bar plots show the experiment results for merging stages, the bottom bar plots show the probabilities of merging edges.

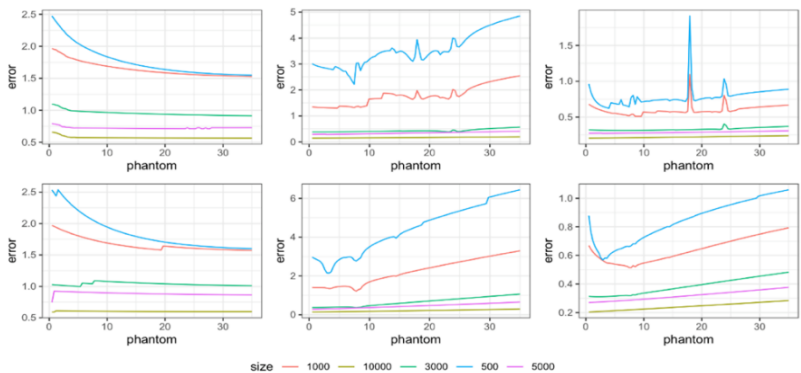

Figure 4. Plots of errors. The three plots on the top are errors for intervened model; the three plots on the bottom are errors (Shenvi & Smith 2018) for the non-intervened model; the left two plots show the result of situational errors the two plots in the middle show the cluster errors; the two plots on the right are average errors between posterior estimation of mean holding time and ground truth.

Thus the explicit incorporation of information extracted from natural language texts through our bespoke causal algebras vastly improves a statistical analysis, especially when informed by large samples. Estimates of conditional probabilities and mean holding times and failure predictions are all then much more accurate once textual information is used.

## 5 Discussion

This short paper outlines some of our recent work in creating causal algebras for remedial intervention. A full exposition, where we explain in detail the hierarchical Bayesian model for structure learning, the causal inference and our elicitation methodology will be reported soon. Our next task is to scale up this methodology so that it become a feasibly implementable tool for larger logs where robust automatic natural language extraction is needed. There is much yet to do. However we hope that we have demonstrated the promise of harnessing free text log data for a reliability analysis by devising a new graphically supported bespoke causal analysis.

**Acknowledgement**

This project is funded by the Engineering and Physical Sciences Research Council (EPSRC) and the statistics department of the University of Warwick. Professor Jim Q. Smith is supported by the Alan Turing Institute and EPSRC with grant number EP/K03 9628/.